\documentclass{bmcart}

\usepackage{amsthm,amsmath,amssymb}
\usepackage[utf8]{inputenc} 

\usepackage{multirow}
\usepackage{booktabs}
\usepackage{graphics}
\usepackage{adjustbox}
\usepackage{subcaption}
\usepackage[font=small,labelfont=bf]{caption}
\usepackage{url}
\usepackage{graphicx}

\startlocaldefs
\endlocaldefs

\begin{document}

\begin{frontmatter}

\begin{fmbox}
\dochead{Research}


\title{Assessment Framework for Deepfake Detection in Real-world Situations}


\author[
  addressref={aff1},                   
  corref={aff1},                       
  email={yuhang.lu@epfl.ch}   
]{\inits{Y.L.}\fnm{Yuhang} \snm{Lu}}
\author[
  addressref={aff1},
  email={touradj.ebrahimi@epfl.ch}
]{\inits{T.E.}\fnm{Touradj} \snm{Ebrahimi}}


\address[id=aff1]{
  \orgdiv{Multimedia Signal Processing Group (MMSPG)},             
  \orgname{\'Ecole Polytechnique F\'ed\'erale de Lausanne (EPFL)},          
  \city{Lausanne},                              
  \cny{Switzerland}                                    
}



\end{fmbox}


\begin{abstractbox}

\begin{abstract} 


Detecting digital face manipulation in images and video has attracted extensive attention due to the potential risk to public trust. To counteract the malicious usage of such techniques, deep learning-based deepfake detection methods have been employed and have exhibited remarkable performance. However, the performance of such detectors is often assessed on related benchmarks that hardly reflect real-world situations. For example, the impact of various image and video processing operations and typical workflow distortions on detection accuracy has not been systematically measured. 
In this paper, a more reliable assessment framework is proposed to evaluate the performance of learning-based deepfake detectors in more realistic settings. To the best of our acknowledgment, it is the first systematic assessment approach for deepfake detectors that not only reports the general performance under real-world conditions but also quantitatively measures their robustness toward different processing operations. 
To demonstrate the effectiveness and usage of the framework, extensive experiments and detailed analysis of three popular deepfake detection methods are further presented in this paper. 
In addition, a stochastic degradation-based data augmentation method driven by realistic processing operations is designed, which significantly improves the robustness of deepfake detectors.





\end{abstract}




\end{abstractbox}
%

\end{frontmatter}




\section{Introduction}

In recent years, the rapid development of deep convolutional neural networks (DCNNs) and ease of access to large-scale datasets have led to significant progress on a broad range of computer vision tasks and meanwhile created a surge of new applications. For example, the recent advancement of generative adversarial networks (GANs) \cite{karras2017progressive, karras2019style, karras2020analyzing} has made it possible to generate realistic forged contents that are difficult for humans to distinguish from their authentic counterparts. In particular, current deep learning-based face manipulation techniques \cite{roessler2019faceforensicspp, thies2019deferred, nirkin2019fsgan,zakharov2019few} are capable of changing the expression, attributes, and even identity of a human face image, the outcome of which refers to the popular term `Deepfake'. The recent development of such technologies and the wide availability of open-source software has simplified the creation of deepfakes, increasingly damaging our trust in online media and raising serious public concerns. To counteract the misuse of these deepfake techniques and malicious attacks, detecting manipulations in facial images and video has become a hot topic in the media forensics community and has received increasing attention from both academia and businesses.

Nowadays, multiple grand challenges, competitions, and public benchmarks \cite{DFDC2020,jiang2020deeperforensics10,chen2022ai} are organized to assist the progress of deepfake detection. At the same time, with the advanced deep learning techniques and large-scale datasets, numerous detection methods \cite{roessler2019faceforensicspp, Nguyen2019UseOA, Zhao2021MultiattentionalDD, liu2021spatial, qian2020thinking, li2021frequency, luo2021generalizing} have been published and have reported promising results on different datasets. But some studies \cite{khodabakhsh2018fake,xuan2019generalization} have shown that the detection performance significantly drops in the cross-dataset scenario, where the fake samples are forged by other unknown manipulation methods. Therefore, cross-dataset evaluation has become an important step in recent studies to better show the advantages of deepfake detection methods, encouraging researchers \cite{haliassos2021lips, kim2021fretal, Shiohara_2022_CVPR} to propose detection methods with better generalization ability to different types of manipulations. 

Nevertheless, another scenario that commonly exists in the real world has received little attention from researchers. In fact, it has long been shown that DCNN-based methods are vulnerable to real-world perturbations and processing operations \cite{Dodge2016UnderstandingHI, mehdipour2016comprehensive, grm2018strengths} in different vision tasks. In more realistic conditions, images and video can face unpredictable distortions from the extrinsic environment, such as noise and poor illumination conditions, or constantly undergo various processing operations to ease their distribution. In the context of this paper, a deployed deepfake detector could mistakenly block a pristine yet heavily compressed image. On the other hand, a malicious agent could also fool the detector by simply adding imperceptible noise to fake media content. To the best of our acknowledgment, most of the current deep learning-based deepfake detection methods are developed based on constrained and less realistic face manipulation datasets and therefore, they are not robust enough in real-world situations. 
Similarly, the conventional assessment approach, which exists in various benchmarks, often directly samples test data from the same distribution as training data and can hardly reflect model performance in more complex situations. In fact, most of the existing deepfake detection methods only report their performance on some well-known benchmarks in the community.

Therefore, a more reliable and systematic approach is desired firsthand in order to assess the performance of a deepfake detector in more realistic scenarios and further motivate researchers to develop robust detection methods. In this paper, a comprehensive assessment framework for deepfake detection in real-world situations has been conceived for both image and video deepfakes. At the same time, a generic approach to improve the robustness of the detectors has been proposed. 

In summary, the following contributions have been made.
\begin{itemize}
    \item A realistic assessment framework is proposed to evaluate and benchmark the performance of learning-based deepfake detection systems. To the best of our knowledge, this is the first framework that systematically evaluates deepfake detectors in realistic situations. 
    \item The performance of several popular deepfake detection methods has been evaluated and analyzed with the proposed performance evaluation framework. The extensive results demonstrate the necessity and effectiveness of the assessment approach.
    \item Inspired by the real-world data degradation process, a stochastic degradation-based augmentation (SDAug) method driven by typical image and video processing operations is designed for deepfake detection tasks. It brings remarkable improvement in the robustness of different detectors.  
    \item A flexible Python toolbox is developed and the source code of the proposed assessment framework is released to facilitate relevant research activities.

\end{itemize}

This article is an extended version of our recent publication \cite{lu2022novel}. The additional contents of this submission are summarized as follows.
\begin{itemize}
    \item More recent deepfake detection methods have been summarized and introduced in the related work section.
    \item The proposed assessment framework has been extended to support the evaluation of video deepfake detectors.  
    \item The performance of a current state-of-the-art video deepfake detection method has been additionally evaluated using the assessment framework.
    \item More substantial experimental results have been presented to better demonstrate the necessity and usage of the assessment framework. The performance and characteristics of three popular deepfake detection methods are analyzed in depth based on the assessment results.  
    \item The impact of different image compression operations on the performance of deepfake detectors is additionally studied in detail.
    \item More experiments and comparisons have been conducted for the proposed stochastic degradation-based augmentation method. Its effectiveness and limitations are further analyzed. 
\end{itemize}

\section{Related Work}

\subsection{Deepfake Detection} 
Deepfake detection is often treated as a binary classification problem in computer vision. Early on, solutions based on facial expressions \cite{Agarwal_2019_CVPR_Workshops}, head movements \cite{Yang2019ExposingDF} and eye blinking \cite{9072088} were proposed to address such detection problems. In recent years, the primary solution to this problem is by leveraging advanced neural network architectures. Zhou et al. \cite{zhou_two-stream_2017} proposed to detect deepfakes with a two-stream neural network. R\"ossler et al. \cite{roessler2019faceforensicspp} retrained an XceptionNet \cite{Chollet2017XceptionDL} with manipulated face dataset which outperforms their proposed benchmark. Nguyen et al. \cite{Nguyen2019UseOA} combined traditional CNN and Capsule networks \cite{sabour2017dynamic}, which require fewer parameters. Some video deepfake detectors \cite{guera2018deepfake, sabir2019recurrent, masi2020two} leveraged recurrent convolutional neural networks to track forgery clues from the temporal sequences. Other creative attempts in network architectures include but are not limited to, multi-task autoencoders \cite{nguyen2019multi, du2020towards}, efficient networks \cite{Montserrat_2020_CVPR_Workshops, Shiohara_2022_CVPR} and vision transformers \cite{zheng2021exploring}.
In addition, the attention mechanism, a well-known technique to highlight the informative regions, has also been applied to further improve the training process of the detection system. Dang et.al \cite{dang2020detection} proposed a detection system based on an attention mechanism. Zhao et al. \cite{Zhao2021MultiattentionalDD} designed multi-attention heads to predict multiple spatial attention maps. Their proposed attention map can be easily implemented and inserted into existing backbone networks. 
Besides focusing on the spatial domain, recent work \cite{liu2021spatial, qian2020thinking, li2021frequency, luo2021generalizing, saikia2021improving} attempts to resolve the problem in the frequency domain. The theory behind them is based on the fact that current popular GAN-based image manipulation methods often introduce low-frequency clues due to the built-in up-sampling operation. These methods transform the image to the frequency domain via DCT transformation and separate information according to different frequency bands. As a result, the forgery traces are more effectively captured.

To tackle the generalization problem, one important branch of work directly trains models with fully synthetic data, which forces the models to learn more generic representations for deepfake detection. For example, Xray \cite{Li2020FaceXF} and SBIs \cite{Shiohara_2022_CVPR} methods manually generate blended faces during the training process as fake samples, which reproduce the blending artifacts existing in real-world GAN-synthesized deepfakes. Both methods have achieved remarkable performance and notable generalization ability to certain types of manipulation methods. But as explained by the authors, these methods are susceptible to many common perturbations, such as low-resolution and heavy compression. 
In this paper, three different types of deepfake detectors \cite{roessler2019faceforensicspp, Nguyen2019UseOA, Shiohara_2022_CVPR} are adopted for experiments. 

\begin{table}[t]
  \centering
  \caption{Deepfake Detection Challenge (DFDC) \cite{DFDC2020} top-5 prize winners and their corresponding results.}
    \begin{tabular}{c|c}
    Team name & Overall log loss \\
    \midrule
    Selim Seferbekov \cite{seferbekov} & 0.4279 \\
    WM \cite{WM}   & 0.4284 \\
    NTechLab \cite{NTech} & 0.4345 \\
    Eighteen Years Old \cite{Eighteen} & 0.4347 \\
    The Medics \cite{Medics} & 0.4371 \\
    \end{tabular}%
  \label{tab:dfdc}%
\end{table}%

\subsection{Deepfake Detection Competitions Review}
To assist in faster progress and better advancement of deepfake detection tasks, numerous large-scale benchmarks, competitions, and challenges \cite{roessler2019faceforensicspp, jiang2020deeperforensics10, DFDC2020, chen2022ai} have been organized, the results of which have been made publicly available. Meta partnered with some academic experts and industry leaders and created the Deepfake Detection Challenge (DFDC) \cite{DFDC2020} in 2019. The competition provided a large incentive, i.e. 1 million USD, for experts in computer vision and deepfake detection to dedicate time and computational resources to train models for benchmarking. More recently, the Trusted Media Challenge (TMC) \cite{chen2022ai} was organized by AI Singapore with a total prize pool of up to 500k USD in order to explore how artificial intelligence technologies could be leveraged to combat fake media. Nevertheless, after a thorough investigation of the benchmarking results, a new question emerges: \textit{Can the assessment approach adopted by the competitions reflect their performance in realistic scenarios?} Although both challenges tried to simulate real-world conditions by preprocessing part of the testing data with some common video processing techniques, they do not really differentiate the detectors. 
As shown in Table \ref{tab:dfdc}, the final results of the top-5 prize winners from DFDC \cite{DFDC2020} are extremely close and the ranking seems to be easily affected by some random noise, for example simply taking out a few fake samples or adding slightly more severe blurriness effect. 

The current ranking approach in these competitions is not reliable. A more rigorous framework is introduced in this work, which is able to differentiate the detectors in multiple dimensions, i.e. general performance, general robustness in realistic conditions, and robustness to specific impacting factors. 

\subsection{Robustness Benchmark}
In recent years, research has been conducted to explore the robustness of CNN-based methods toward real-world image corruption. Dodge and Karam \cite{Dodge2016UnderstandingHI} measured the performance of image classification models with data disturbed by noise, blurring, and contrast changes. In \cite{hendrycks2019robustness}, Hendrycks et al. presented a corrupted version of ImageNet \cite{5206848} to benchmark the robustness of image recognition models against common image manipulations. \cite{michaelis2019dragon, 2020, Sakaridis_2021_ICCV} focused on a safety-critical task, autonomous driving, and provided a robustness benchmark for various relevant vision tasks, such as object detection and semantic segmentation. Similar work has been done for face recognition tasks, \cite{karahan2016image,mehdipour2016comprehensive,grm2018strengths,9922840} analyzed the robustness of CNN-based face recognition models towards face variations caused by illumination change, occlusion, and standard image processing operations. In the media forensics community, StirMark \cite{petitcolas1998attacks} tested the robustness of image watermarking algorithms. The ALASKA\#2 dataset \cite{cogranne2020alaska} was created following a careful evaluation of ISO parameters, JPEG compression, and noise level on FlickR images, etc., in order to help researchers in designing way more general and robust steganographic and steganalysis methods. 
It is worth noting that two popular deepfake detection benchmarks, DFDC \cite{DFDC2020} and Deeperforensics-1.0 \cite{jiang2020deeperforensics10} also adopted standard processing operations to part of the testing data. They randomly applied distortions to a small portion of test data and considered only one severity level for each processing operation. However, the way they evaluate a detector's robustness is not systematic enough. The assessment results cannot rigorously show to which extent the detector is affected by the distorted data, nor help identify which factors show more significant influence on the detector's performance. 
There is a lack of a fair and flexible methodology that systematically compares the performance of deepfake detectors in realistic situations. In this work, a new assessment framework is introduced to solve this problem.

\newcommand\w{0.22\linewidth}
\newcommand\y{\linewidth}
\begin{figure}[t]
\centering
\begin{subfigure}[b]{\w}
  \includegraphics[width=\y]{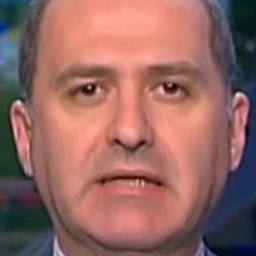}  
  \caption{Unaltered}
\end{subfigure}%
\hfill
\begin{subfigure}[b]{\w}
  \includegraphics[width=\y]{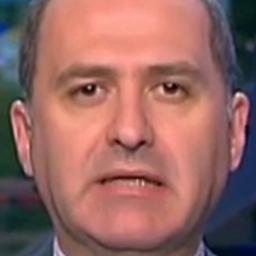}  
  \caption{JPEG}
\end{subfigure}%
\hfill
\begin{subfigure}[b]{\w}
  \includegraphics[width=\y]{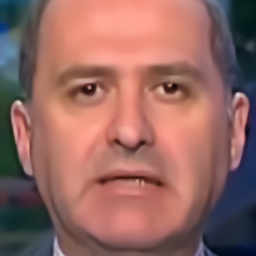}  
  \caption{DL-Comp}
\end{subfigure}%
\hfill
\begin{subfigure}[b]{\w}
  \includegraphics[width=\y]{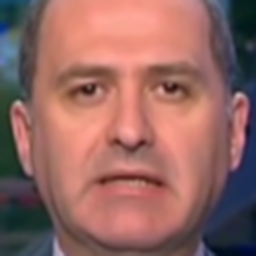}  
  \caption{GB}
\end{subfigure}%
\hfill
\begin{subfigure}[b]{\w}
  \includegraphics[width=\y]{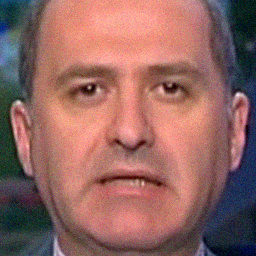}  
  \caption{GN}
\end{subfigure}%
\hfill
\begin{subfigure}[b]{\w}
  \includegraphics[width=\y]{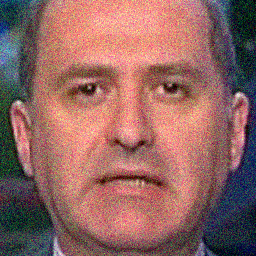}
  \caption{Po-Gau-Noise}
\end{subfigure}%
\hfill
\begin{subfigure}[b]{\w}
  \includegraphics[width=\y]{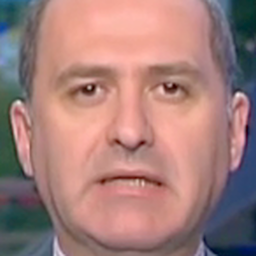}  
  \caption{GammaCorr}
\end{subfigure}%
\hfill
\begin{subfigure}[b]{\w}
  \includegraphics[width=\y]{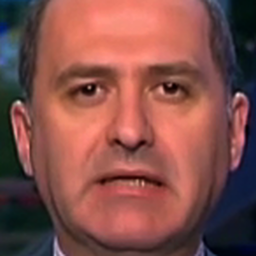}
  \caption{GammaCorr}
\end{subfigure}%
\hfill
\begin{subfigure}[b]{\w}
  \includegraphics[width=\y]{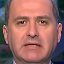}  
  \caption{Resize}
\end{subfigure}%
\hfill
\begin{subfigure}[b]{\w}
  \includegraphics[width=\y]{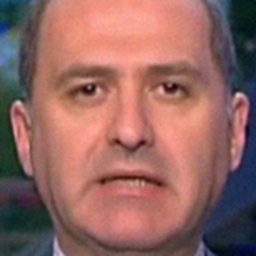}  
  \caption{GN+GB}
\end{subfigure}%
\hfill
\begin{subfigure}[b]{\w}
  \includegraphics[width=\y]{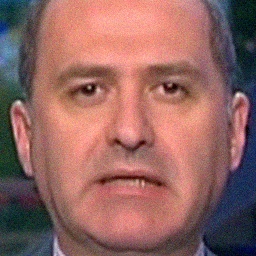}  
  \caption{JPEG+GN}
\end{subfigure}%
\hfill
\begin{subfigure}[b]{\w}
  \includegraphics[width=\y]{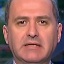}  
  \caption{JPEG+Resize}
\end{subfigure}
\caption{Example of a typical image in the FFpp test set after applying various image processing operations. Some notations are explained as follows. DL-Comp: Deep learning-based compression. GB: Gaussian blur. GN: Gaussian noise. Po-Gau-Noise: Poissonian-Gaussian noise. GammaCorr: Gamma correction. Resize: Reduce resolution. $+$: Combination of two operations.}
\label{fig:image-op-example}
\end{figure}


\section{Proposed Assessment Framework}
Nowadays, deepfakes are distributed on the internet in both image and video formats. Some of the detection methods are targeted for both cases, while others are specially designed for video deepfake detection. The proposed assessment framework is designed in a way that the performance of a deepfake detection can be evaluated under either image or video scenario. 

In this section, the common realistic influencing factors for image and video deepfakes are first introduced respectively. 
Then, the proposed assessment framework is described in order to provide a fair comparison for deepfake detectors under more realistic situations.

\subsection{Realistic Influencing Factors for Image Deepfakes}

In a real-world situation, the images are often processed by various digital image processing operations before being distributed. 
In more adverse cases, malicious deepfakes can be slightly corrupted to fool the detector while maintaining good perceptual quality. It is still unknown to which extent the popular deepfake detectors are able to make correct predictions.
In this context, the most prominent factors have been considered in the assessment framework. 

In general, the framework contains six categories of image processing operations or corruptions with more than ten minor types. Each type consists of multiple severity levels. The details of all operations used in evaluations are described below with the illustration of a typical example in Figure \ref{fig:image-op-example}. Specifically, the following factors are considered in the assessment framework.

\textbf{Noise}: Noise is a typical distortion especially when images are captured in a low illumination condition. To simulate the noise, an Additive White Gaussian Noise (AWGN) is applied to the data and the pixel values are clipped to [0, 255]. In this paper, the variance value $\sigma$ is selected in a range from 5 to 50. In addition, Poissonian-Gaussian noise \cite{4623175} is also included to better reflect the realistic noise levels, whose parameters are learned from a group of real noisy pictures. 

\textbf{Resizing}: Resizing is one of the most commonly used image processing operations. It refers to changing the dimensions of the media content to fit the display or other purposes. On the other hand, the resizing operation, more specifically the down-sampling operation can significantly reduce the performance of modern deep learning-based detectors \cite{article, Li2019OnLF} due to a loss of discriminative information. This is often the case for those earlier image contents that are of poor quality.
In this framework, the impact of resizing operation is simulated by first downscaling the images and then upscaling back using bicubic interpolation. 

\textbf{Image Compression}: Lossy compression refers to the class of data encoding methods that remove unnecessary or less important information and only use partial data to represent the content. These techniques are used to reduce data size for efficient storage and transmission of content and are widely applied to image processing. In this framework, the JPEG compression artifacts are applied and the impact of different quality factors, i.e. from 30 to 95, on the deepfake detection system is evaluated. As deep learning-based compression techniques are becoming increasingly popular in this community, two AI-based image compression techniques \cite{balle2018variational, mentzer2020high} are also considered in this framework with multiple compression qualities to choose from. 

\textbf{Denoising}: A typical way to reduce noise is by smoothing, which is a low-pass filtering applied to the image. 
The denoising operation is often applied to image contents after being acquired by the camera
but at the same time, it tends to blur the media content and results in a reduction of details, which is harmful to the detection system. To measure the impact of the denoising operation, the blurriness effect is simulated in our framework by applying Gaussian filters with kernel size $\sigma$ ranging from 3 to 11. Meanwhile, learning-based denoising techniques are gradually deployed in practice. They recover a noisy image with higher quality but often bring unpredictable artifacts. The impact of applying the DnCNN technique \cite{7839189} is assessed in the framework.

\textbf{Enhancement}: In realistic conditions, the image data captured in the wild can suffer from poor illumination. Image enhancement is frequently used to adjust the media content for better display. In this assessment framework, the contrast and brightness of the test data are modified by both linear and nonlinear adjustments. The former simply adds or reduces a constant pixel value while the latter applies gamma correction. 

\textbf{Combinations}: It is even more common that the media content suffers from multiple types of distortions and processing operations. Therefore, the mixture of two or three operations above is also considered, such as combining JPEG compression and Gaussian noise, making the test data better reflect more complex real-world scenarios.

\begin{figure}[t]
\centering
\begin{subfigure}[b]{\w}
  \includegraphics[width=\y]{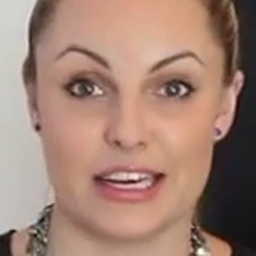}  
  \caption{Unaltered}
\end{subfigure}%
\hfill
\begin{subfigure}[b]{\w}
  \includegraphics[width=\y]{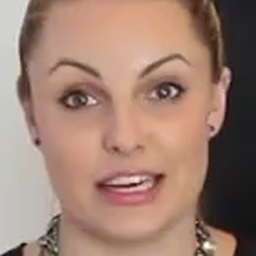}  
  \caption{C23}
\end{subfigure}%
\hfill
\begin{subfigure}[b]{\w}
  \includegraphics[width=\y]{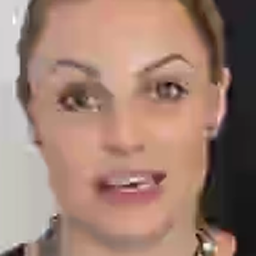}  
  \caption{C40}
\end{subfigure}%
\hfill
\begin{subfigure}[b]{\w}
  \includegraphics[width=\y]{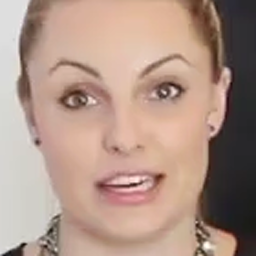}  
  \caption{Light}
\end{subfigure}%
\hfill
\begin{subfigure}[b]{\w}
  \includegraphics[width=\y]{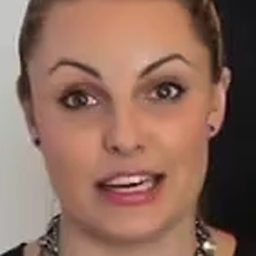}  
  \caption{Dark}
\end{subfigure}%
\hfill
\begin{subfigure}[b]{\w}
  \includegraphics[width=\y]{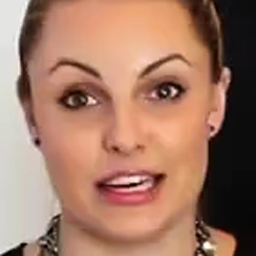}
  \caption{Contrast}
\end{subfigure}%
\hfill
\begin{subfigure}[b]{\w}
  \includegraphics[width=\y]{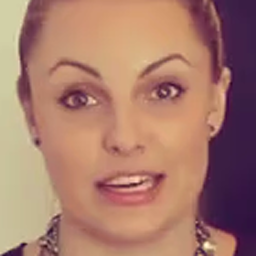}
  \caption{Vintage}
\end{subfigure}%
\hfill
\begin{subfigure}[b]{\w}
  \includegraphics[width=\y]{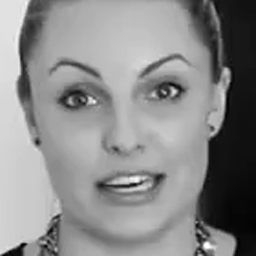}  
  \caption{Grayscale}
\end{subfigure}%
\hfill
\begin{subfigure}[b]{\w}
  \includegraphics[width=\y]{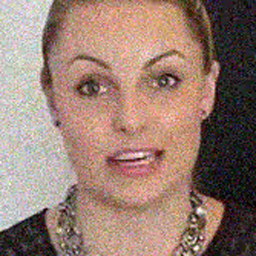}  
  \caption{Gaussian Noise}
\end{subfigure}%
\hfill
\begin{subfigure}[b]{\w}
  \includegraphics[width=\y]{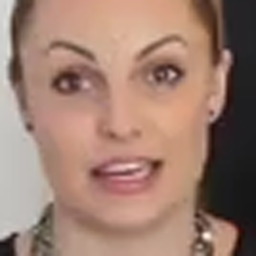}  
  \caption{Resolution}
\end{subfigure}%
\hfill
\begin{subfigure}[b]{\w}
  \includegraphics[width=\y]{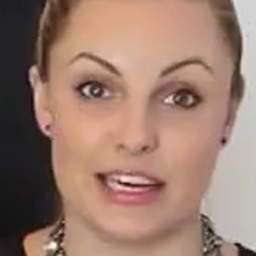} 
  \caption{Hflip}
\end{subfigure}%
\hfill
\begin{subfigure}[b]{\w}
  \includegraphics[width=\y]{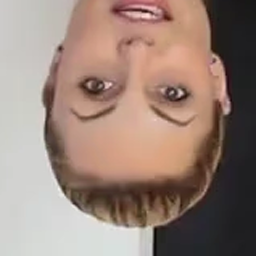}  
  \caption{Vflip}
\end{subfigure}
\caption{Example of a typical video frame in the FFpp test set after applying different video processing operations. Some notations are explained as follows. C23 and C40: Video compression using H.264 codec with factors of 23 and 40. Light and Dark: Increase and decrease brightness. Resolution: Reduce video resolution. Hflip and Vflip: Horizontal and Vertical flip.}
\label{fig:video-op-example}
\end{figure}

\subsection{Realistic Influencing Factors for Video Deepfakes}
Face forgeries by deepfake technology are spread over the Internet not only in the form of images but also as video. The processing operations and various video effects are very common on different social media, smartphone applications, and streaming platforms. Their impact on the accuracy of detection methods should not be neglected. 

The framework includes seven categories of video processing operations with commonly used parameters. The illustrative example of testing data is shown in Figure~\ref{fig:video-op-example}. The factors are also described in detail as follows. 


\textbf{Video Compression:}
Similar to images, uncompressed raw video requires a large amount of storage space. Although lossless video compression codecs can perform at a compression factor of 5 to 12, a typical lossy compression video can achieve a much lower data rate while maintaining high visual quality. In fact, compression technologies for video provide the basis for the distribution of video worldwide.
The potential deepfake video propagates among social networks after being compressed several times. However, the possible side effect of lossy compression artifacts on deep learning-based detectors has not been sufficiently studied. It is necessary to test the robustness of a deepfake detector on compressed authentic and deepfake video. In this context, the proposed assessment framework consists of test data compressed by H.264 codec using the FFMPEG toolbox with two constant rate factors, namely 23 and 40. 

\textbf{Flip:}
Flipping a video horizontally describes the creation of a mirror video of the original footage. It is a very common video editing method that prevents video cuts from disorienting the viewer. But whether and to which extent the flipping operation can affect a deepfake detector has not been evaluated before. 
On the other hand, the vertical flipping operation is one of the easiest ways to fool a detector. In fact, most current detectors will not adjust or correct the face pose during preprocessing step. Hence, one can simply upload a flipped video to avoid being detected while it is still readable to a human. 

\textbf{Video Filters:}
In recent years, video filters have become popular on social media. They are preset treatments included in many video editing apps, software, and social media platforms, providing easy access for users to alter the look of a video clip. Some common types of video filters include color filters, beauty filters, stylization filters, etc. 
The overall color palette of a deepfake video can be changed by a video filter on social media, making it an out-of-distribution sample from common deepfake databases. 
In the proposed assessment framework, two typical filters, `Vintage' and `Grayscale', are considered.

\textbf{Brightness:}
Brightness is a measure of the overall lightness or darkness of a video. 
Adjusting the brightness of a video can affect the way that colors are perceived, as well as the visibility of details and textures. For example, increasing the brightness can make it easier to see details in shadows, while decreasing the brightness can obscure details in highlights. 
In real-world conditions, the brightness of a video is often adjusted to create a different sense of style of a video. 
The assessment framework takes this situation into consideration and measures the performance of a detector under different brightness conditions. More specifically, the `Lighten' and `Darken' commands in the FFMPEG toolbox are applied to the testing video respectively.

\textbf{Contrast:}
Contrast refers to the difference between the lightest and darkest areas of a video. Similar to brightness, adjusting contrast is one of the most common operations to change the visual appearance of a video. The `Contrast' command in the FFMPEG toolbox is employed to increase the contrast of the testing video. 

\textbf{Noise:}
Similar to images, video noise is a common problem in video clips shot in low-light conditions or with small-sensor devices such as mobile phones. It often appears as annoying grains and artifacts in the video. Gaussian noise with a temporal variance but fixed strength is applied to the video data. 

\textbf{Resolution:}
Resolution refers to the number of pixels in a video. There is an important trade-off between the resolution and file size. Decreasing the resolution of a video will generally result in a lower-quality video with fewer details to be displayed on the screen. But it can also reduce the file size, which makes it easier to store and share. On the other hand, the resolution change can also affect the ratio of the width to the height of the video. The performance of the deepfake detector when facing low-resolution or stretched video will be evaluated by the proposed framework.


\begin{figure*}[ht]
	\centering
	\begin{adjustbox}{width=0.95\textwidth}
    \includegraphics[]{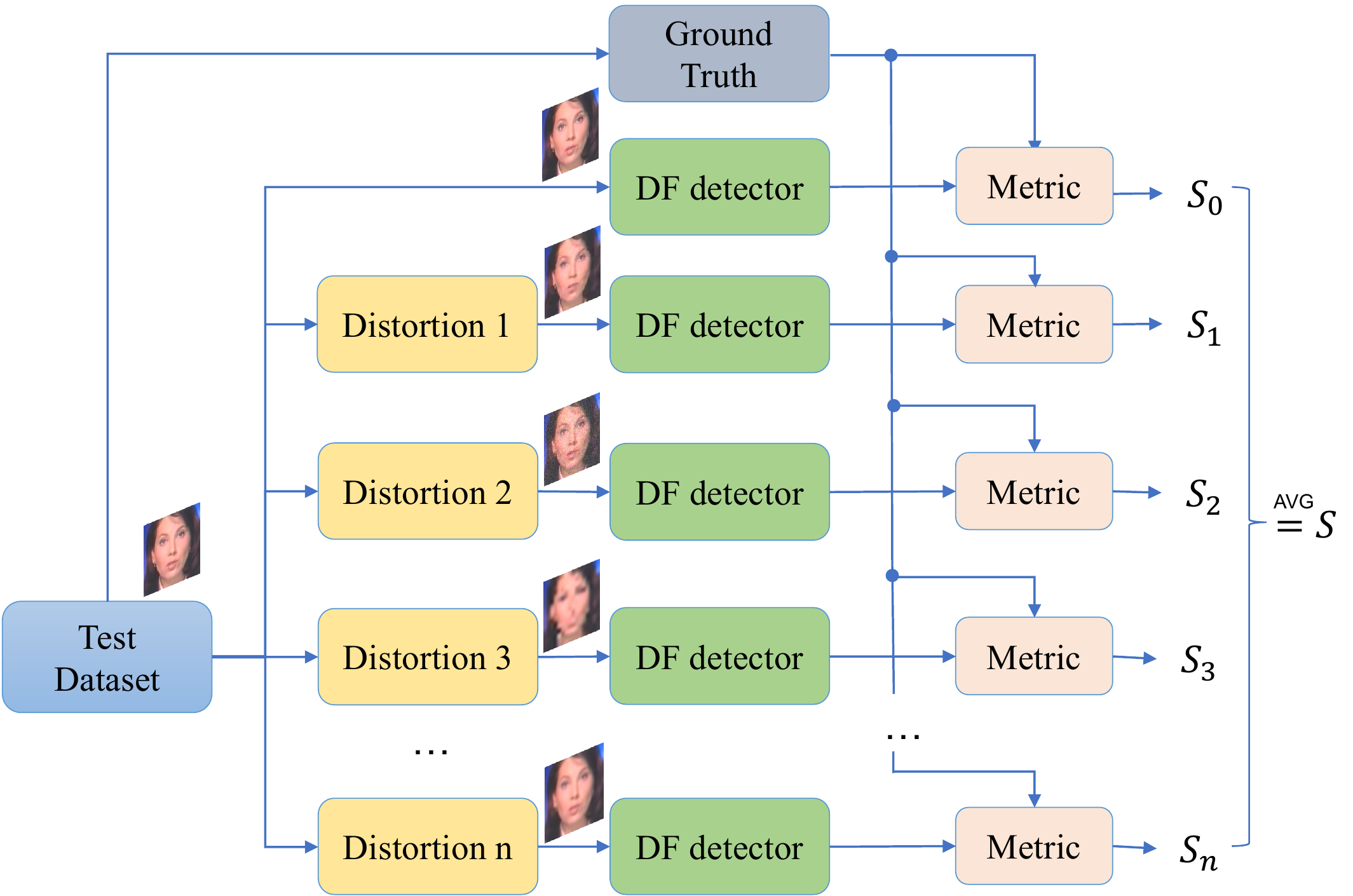}
	\end{adjustbox}
	\caption{Workflow of the proposed assessment framework.}
	\label{fig:af}
\end{figure*}


\subsection{Assessment Methodology}

Current deepfake detection algorithms are based on deep learning and rely heavily on the distribution of the training data. 
These methods are typically evaluated using a test dataset that is similar to the training sets. Some benchmarks, such as \cite{DFDC2020, jiang2020deeperforensics10}, attempt to measure the performance of deepfake detectors under more realistic conditions by adding random perturbations to partial test data and mixing up with others. However, there is no standard approach for determining the proportion or strength of these perturbations, which makes the results of these benchmarks more stochastic and less reliable. The assessment methodology proposed in this paper aims to more thoroughly measure the impact of various influencing factors, at different severity levels, on the performance of deepfake detection algorithms.

In this section, the principle and usage of our assessment framework are introduced in detail. First, the deepfake detector is trained on its original target datasets, such as FaceForensics++ \cite{roessler2019faceforensicspp}. The processing operations and corruptions in the framework are not applied to the training data. Then, as illustrated in Figure \ref{fig:af}, multiple copies of the test set are created, and each type of distortion at one specific severity level is applied to the copies independently. The standard test data together with different distorted data are fed to the deepfake detector respectively. Finally, the detector generates real or fake predictions and calculates performance metrics for each processed dataset. 
An overall evaluation score is obtained by averaging the scores from each distortion style and strength level to report the general performance of a tested detector.  
Besides, the computed metrics can also be grouped by each operation category to further analyze the robustness of one deepfake detector on a specific processing operation. 

In addition, in order to relieve the burden on storage caused by the multiple copies of the test set, a Python toolbox is developed to address this problem in an online manner, which hard-codes the digital processing operations and makes the strength level a parameter. It operates in the same format as the famous Transforms module in the TorchVison toolbox and can be easily integrated into the evaluation process.

\section{Assessment Results}
In this work, numerous experiments have been conducted to demonstrate the effectiveness and usage of the proposed assessment framework. The experimental setup will be described at the beginning of this section, followed by the substantial assessment results and analysis for both image and video scenarios. In the end, the impact of three image compression technologies on deepfake detectors is further discussed as an example of the multiple applications of the framework. 

\subsection{Implementation Details}

\subsubsection{Datasets}
Two widely used face manipulation datasets are selected in this paper for extensive experimentation.

\textbf{FaceForensics++} \cite{roessler2019faceforensicspp}, denoted by FFpp, contains 1000 pristine and 4000 manipulated video generated by four different deepfake creation algorithms. Additionally, raw video contents are compressed with two quality parameters using the AVC/H.264 codec, denoted as C23 and C40. In the experiments, the training set is denoted as \textit{FFpp-Raw}, \textit{FFpp-C23}, and \textit{FFpp-C40} when the model is trained on single-quality-level data, while it is denoted as \textit{FFpp-Full} when data of all three quality levels are involved for training. On the contrary, to provide a fair baseline, only uncompressed data are used for the final assessment.  

\textbf{Celeb-DFv2} \cite{Celeb_DF_cvpr20} is another high-quality dataset, with 590 pristine celebrity video and 5639 fake video. The test data is selected as recommended by \cite{Celeb_DF_cvpr20} while the rest are left for training purposes. In our experiment, the training and validation set was split into 80\% and 20\% accordingly.

\subsubsection{Detection Methods}
Experiments have been conducted with the following learning-based deepfake detectors, all of which have reported excellent performance on popular benchmarks.  

\textbf{Capsule-Forensics} is a deepfake detection method based on a combination of capsule networks and CNNs. The capsule network was initially proposed by \cite{sabour2017dynamic} to address some limitations of CNNs and it used a rather smaller amount of parameters than traditional CNN to train very deep neural networks. \cite{Nguyen2019UseOA} employed the capsule network as a component in a deepfake detection pipeline for detecting manipulated images and video. This method achieved the best performance at that time in the FaceForensics++ dataset compared to its competing methods.

\textbf{XceptionNet} \cite{Chollet2017XceptionDL} is a popular CNN architecture in many computer vision tasks and has been used to detect face manipulations when it works as a classification network. Ro\"ssler et al. \cite{roessler2019faceforensicspp} first adopted it as a baseline in the FaceForensics++ benchmark along with three other approaches. The detection system based on XceptionNet architecture was first pre-trained using ImageNet database \cite{5206848} and then re-trained on a specific dataset for the deepfake detection task. 
It achieved excellent performance in the FaceForensics++ benchmark on both compressed and uncompressed contents and has become a popular baseline method for recent deepfake detection approaches.

\textbf{SBIs} \cite{Shiohara_2022_CVPR} refers to a data synthetic method, Self-blended Images, which is specially designed for deepfake detection tasks. This method aims to generate hardly recognizable fake samples that contain common face forgery traces in order to encourage the model to learn more general and robust representations for face forgery detection. 
The overall detection system is based on a pre-trained deep classification network, EfficientNet-b4 \cite{tan2019efficientnet}. After retraining with the SBIs technique, the detector demonstrates an impressive generalization ability to different unseen face manipulations and achieves the current state-of-the-art in cross-dataset settings. But its robustness to common image and video processing operations has not been measured. 

\subsubsection{Training Details}
Both the Capsule-Forensics and XceptionNet are trained with Adam optimizer with $\beta_1=0.9$, $\beta_2=0.999$. The Capsule-Forensics model is trained from scratch for 25 epochs with a learning rate of $5 \times 10^{-4}$, and the XceptionNet model is trained for 10 epochs with a learning rate of $1 \times 10^{-3}$. For both methods, 100 frames are randomly sampled from each video for training purposes and 32 frames are extracted for validation and testing. Extracted frames are pre-processed and cropped around the face regions using the dlib toolbox \cite{dlib09}. The face regions are finally resized into 300x300 pixels before feeding to the network.

The SBIs method has a different experimental setting from the previous two methods. It is retrained with SAM \cite{foret2020sharpness} optimizer for 100 epochs. The batch size and learning rate are set to 32 and $1 \times 10^{-3}$ respectively. During the training phase, only authentic high-quality video is used and the corresponding fake samples are created by their proposed self-blending method. Only 8 frames per video are sampled for training while 32 frames are for validation and testing. 

\subsubsection{Performance Metrics}

During the evaluation, the Area Under Receiver Operating Characteristic Curve (AUC) is used as a metric in all experiments.


\begin{table}[t]
  \centering
  \caption{AUC (\%) scores of the Capsule-Forensics, denoted as CapsuleNet, and XceptionNet methods tested on unaltered and distorted variants of FFpp and Celeb-DFv2 test set respectively. Raw, C23, and Full refer to different quality settings of the FFpp. DL-Comp refers to deep learning-based compression \cite{balle2018variational} and `High' refers to the high-quality compressed image.}
  \begin{adjustbox}{width=\textwidth}

    \begin{tabular}{cccccccccccccccc}
    \toprule
    \multirow{2}[4]{*}{Methods} & \multirow{2}[4]{*}{TrainSet} & \multirow{2}[4]{*}{Unaltered} & \multicolumn{4}{c}{JPEG}      & \multicolumn{4}{c}{DL-Comp}   & \multicolumn{4}{c}{Gaussian Noise} & \multirow{2}[4]{*}{\shortstack{Po-Gau\\ Noise}} \\
\cmidrule{4-15}          &       &       & 95    & 60    & 30    & AVG   & High  & Med   & Low   & AVG   & 5     & 10    & 30    & AVG   &  \\
    \midrule
    \midrule
    \multirow{4}[2]{*}{CapsuleNet} & FFpp-Raw & 99.20 & 97.91 & 76.48 & 59.60 & 78.00 & 55.24 & 54.50 & 50.92 & 53.55 & 61.80 & 59.73 & 51.26 & 57.60 & 55.63 \\
          & FFpp-C23 & 96.32 & 95.09 & 95.76 & 74.91 & 88.59 & 56.96 & 57.42 & 81.57 & 65.32 & 84.51 & 78.56 & 58.63 & 73.90 & 70.59 \\
          & FFpp-Full & 94.52 & 94.95 & 92.18 & 84.50 & 90.54 & 86.83 & 60.98 & 55.69 & 67.83 & 89.03 & 78.54 & 57.95 & 75.17 & 64.87 \\
          & CelebDFv2 & 99.14 & 99.32 & 98.88 & 93.07 & 97.09 & 99.01 & 96.77 & 88.95 & 94.91 & 95.24 & 63.27 & 59.20 & 72.57 & 87.06 \\
    \midrule
    \multirow{4}[1]{*}{XceptionNet} & FFpp-Raw & 99.56 & 76.77 & 56.00 & 54.20 & 62.32 & 50.16 & 50.37 & 50.10 & 50.21 & 50.12 & 51.00 & 50.36 & 50.49 & 48.98 \\
          & FFpp-C23 & 98.64 & 98.01 & 95.99 & 82.77 & 92.26 & 96.11 & 56.25 & 55.71 & 69.36 & 71.35 & 55.84 & 50.87 & 59.35 & 51.48 \\
          & FFpp-Full & 99.02 & 99.00 & 94.78 & 87.86 & 93.88 & 94.36 & 54.88 & 55.78 & 68.34 & 59.00 & 55.09 & 54.08 & 56.06 & 51.43 \\
          & CelebDFv2 & 99.73 & 99.78 & 99.59 & 97.76 & 99.04 & 96.23 & 90.23 & 75.46 & 87.31 & 94.85 & 69.87 & 52.50 & 72.41 & 86.87 \\
    \midrule
    \multirow{2}[3]{*}{Methods} & \multirow{2}[3]{*}{TrainSet} & \multicolumn{4}{c}{Gaussian Blur} & \multicolumn{5}{c}{Gamma Correction}  & \multicolumn{4}{c}{Resize}    & \multirow{2}[3]{*}{\shortstack{Overall\\Average}} \\
\cmidrule{3-15}          &       & 3     & 7     & 11    & AVG   & 0.1   & 0.75  & 1.3   & 2.5   & AVG   & x4    & x8    & x16   & AVG   &  \\
    \midrule
    \midrule
    \multirow{4}[2]{*}{CapsuleNet} & FFpp-Raw & 67.19 & 58.22 & 52.26 & 59.22 & 50.50 & 98.86 & 99.17 & 96.12 & 86.16 & 67.48 & 53.18 & 53.10 & 57.92 & 65.41 \\
          & FFpp-C23 & 85.21 & 53.94 & 52.04 & 63.73 & 52.08 & 95.06 & 96.72 & 92.91 & 84.19 & 79.33 & 64.62 & 50.33 & 64.76 & 73.41 \\
          & FFpp-Full & 85.72 & 58.83 & 56.05 & 66.87 & 56.02 & 93.86 & 93.87 & 85.44 & 82.30 & 84.65 & 65.34 & 52.02 & 67.34 & 75.01 \\
          & CelebDFv2 & 99.01 & 91.04 & 77.52 & 89.19 & 79.60 & 98.52 & 99.40 & 94.62 & 93.04 & 89.22 & 66.98 & 61.94 & 72.71 & 86.58 \\
    \midrule
    \multirow{4}[2]{*}{XceptionNet} & FFpp-Raw & 68.76 & 55.61 & 50.70 & 58.36 & 54.66 & 98.66 & 99.57 & 70.45 & 80.84 & 68.60 & 55.80 & 50.45 & 58.28 & 60.08 \\
          & FFpp-C23 & 95.38 & 59.92 & 52.33 & 69.21 & 58.43 & 98.34 & 98.64 & 91.35 & 86.69 & 92.55 & 70.27 & 56.00 & 72.94 & 74.97 \\
          & FFpp-Full & 96.36 & 70.51 & 54.50 & 73.79 & 51.38 & 98.91 & 98.84 & 88.91 & 84.51 & 93.47 & 75.30 & 60.55 & 76.44 & 75.50 \\
          & CelebDFv2 & 98.77 & 91.81 & 79.94 & 90.17 & 74.97 & 99.53 & 99.74 & 97.49 & 92.93 & 96.01 & 72.21 & 63.03 & 77.08 & 86.49 \\
    \bottomrule
    \end{tabular}%
   \end{adjustbox}
  \label{tab:results-image}%
\end{table}%



\begin{figure}[h]
	\centering
    \includegraphics[width=0.9\linewidth]{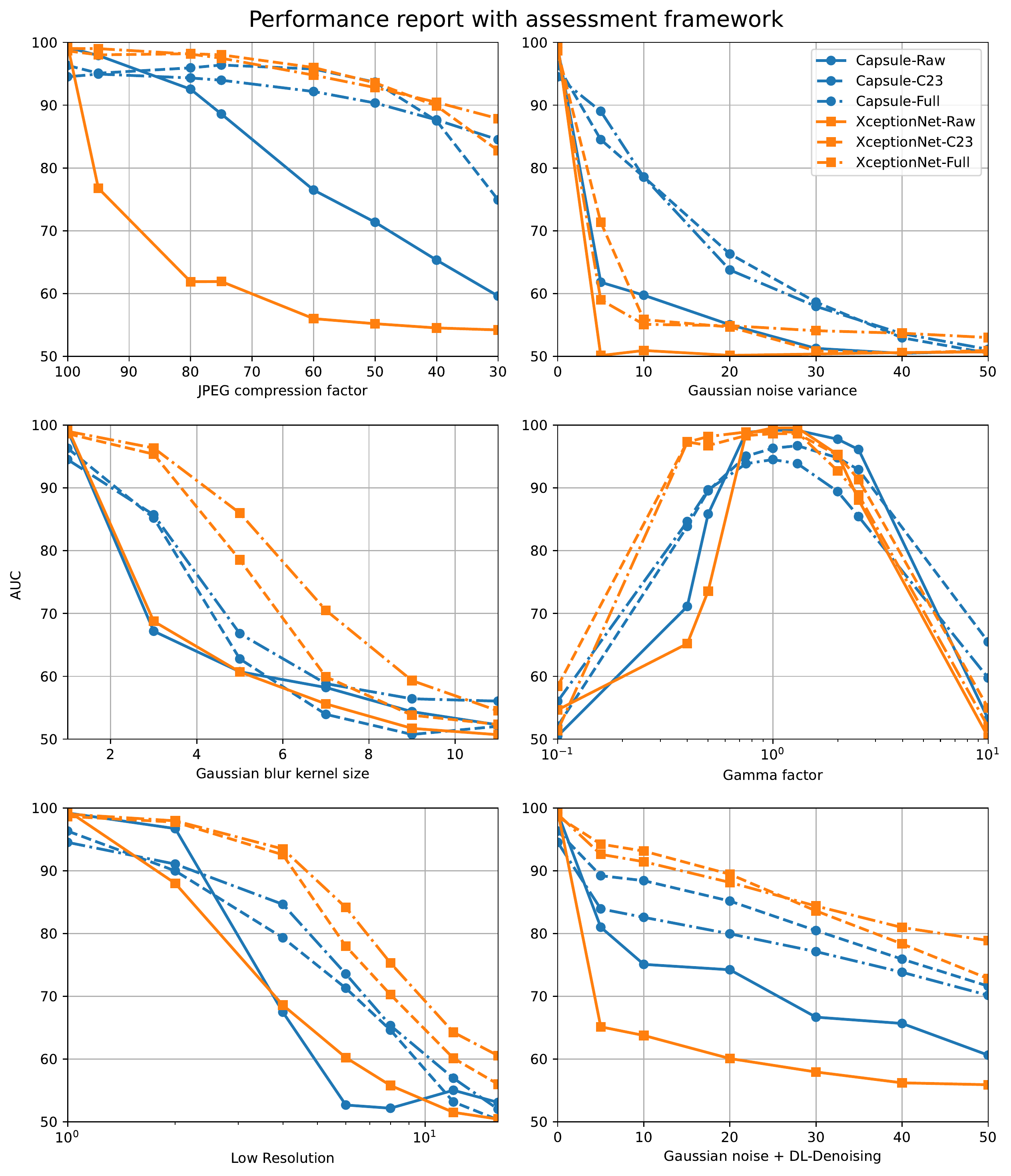}
	\caption{Assessment results of two models trained on FFpp dataset. The suffixes of legends refer to the qualities of the training data. \textit{Full} means using all available quality data for training.}
	\label{fig:results-image}
\end{figure}



\subsection{Assessment Results on Realistic Image Deepfakes}

In this section, the performance of the Capsule-Forensics and XceptionNet methods is measured when facing more realistic image deepfakes produced by the assessment framework. The two deepfake detectors are trained on the original unaltered training sets of FFpp and Celeb-DFv2 respectively. The assessment framework further evaluates the performance of these detectors and summarizes the results as shown in Table \ref{tab:results-image} and Figure \ref{fig:results-image}. 

In general, our findings draw the following conclusions. First, even mild real-world processing operations can have a noticeable negative impact on detection accuracy. The two detectors present exceptional performance on unaltered FFpp and CelebDFv2 testing data as expected, but then show severe performance deterioration on all kinds of modified data from the assessment framework, which indicates a lack of robustness. 

Secondly, the two detectors are prone to be affected by different types of perturbation. When trained on the same high-quality dataset, the Capsule-Forensics method is generally more robust toward JPEG compression, synthetic noise, and gamma correction operation, while XceptionNet at times presents slightly better results that could be of statistical nature. The results from the assessment framework provide valuable guidance toward improving a specific deepfake detector. Moreover, among the considered influencing factors, noise and blurriness effects on images are the most prominent for deepfake detectors. The performance of both detectors deteriorates rapidly after increasing the severity levels of the two distortions. 

Finally, the impact of quality variants of training data on learning-based detectors has been analyzed based on the assessment results. Both the Capsule-Forensics and XceptionNet models trained only with very high-quality data (\textit{FFpp-Raw}) will be extremely sensitive to nearly all kinds of realistic processing operations. 
On the contrary, training the model with relatively low-quality data slightly improves the robustness toward low-intensity processing operations and distortions, but with a cost on the original high-quality testing set. For example, both models trained with compressed data (\textit{FFpp-C23}, \textit{FFpp-Full}) show a higher AUC score on our realistic benchmark, but their performance on original unaltered data decreases by 0.5 - 1\%.


\begin{table}[htbp]
  \centering
  \caption{AUC (\%) scores of three selected deepfake detection methods on the distorted variants of the FFpp test set that are subject to different video processing operations.   
  The notations C23 and C40 here refer to the two different compression rates using AVC/H.264 codec. The notation Resolution refers to reducing video resolution by a specific scale.}
  \begin{adjustbox}{width=\textwidth}
    \begin{tabular}{ccccccccccccccc}
    \toprule
    \multirow{2}[4]{*}{Methods} & \multirow{2}[4]{*}{TrainSet} & \multicolumn{2}{c}{Compression} & \multicolumn{2}{c}{Brightness} & \multirow{2}[4]{*}{Grayscale} & \multirow{2}[4]{*}{Contrast} & \multicolumn{2}{c}{Flipping} & \multicolumn{2}{c}{Resolution} & \multirow{2}[4]{*}{\shortstack{Gaussian \\ Noise}} & \multirow{2}[4]{*}{\shortstack{Vintage \\ Filter}} & \multirow{2}[4]{*}{Average} \\
\cmidrule{3-6}\cmidrule{9-12}          &       & C23 & C40 & Increase & Decrease &       &       & Horizontal & Vertical & x2    & x4    &       &       &  \\
    \midrule
    \midrule
    CapsuleNet & \multirow{3}[2]{*}{FFpp-Raw} & 77.97 & 54.14 & 73.31 & 70.62 & 68.38 & 69.31 & 73.13 & 63.20 & 65.43 & 56.99 & 54.14 & 72.94 & 66.63 \\
    XceptionNet &       & 69.49 & 55.70 & 65.92 & 66.40 & 65.51 & 65.32 & 65.26 & 57.36 & 57.23 & 55.90 & 50.50 & 66.90 & 61.79 \\
    SBIs  &       & 90.43 & 76.27 & 86.38 & 86.47 & 86.27 & 85.94 & 85.98 & 79.28 & 76.35 & 63.62 & 71.52 & 86.54 & 81.25 \\
    \midrule
    CapsuleNet & \multirow{3}[2]{*}{FFpp-C23} & 95.61 & 66.03 & 93.27 & 92.31 & 87.43 & 91.55 & 91.98 & 71.49 & 80.28 & 67.56 & 53.50 & 88.86 & 81.66 \\
    XceptionNet &       & 98.34 & 70.71 & 97.07 & 96.65 & 93.17 & 96.34 & 96.20 & 66.82 & 83.42 & 72.03 & 51.04 & 94.99 & 84.73 \\
    SBIs  &       & 91.71 & 75.43 & 87.63 & 86.51 & 87.31 & 87.40 & 86.84 & 81.22 & 75.40 & 64.31 & 57.06 & 86.28 & 80.59 \\
    \bottomrule
    \end{tabular}%
    \end{adjustbox}
  \label{tab:results-video}%
\end{table}%

\subsection{Assessment Results on Realistic Video Deepfakes}
In addition to images, the framework provides a comprehensive evaluation for the three detection methods, i.e. Capsule-Forensics, XceptionNet, and SBIs, on video deepfakes under real-world conditions. Table \ref{tab:results-video} summarizes the performance of the three deepfake detection methods using the proposed realistic benchmark. 

As a result, when trained with high-quality data, both the Capsule-Forensics and XceptionNet methods show a similar trend as in the previous image deepfake detection benchmark and perform poorly when facing pre-processed video deepfakes. 
Whilst, the SBIs method outperforms the other two detectors and presents a relatively stable score in front of most video processing operations, particularly those artifacts introduced by changing brightness or assigning video filters. 

However, when the previous two methods are trained directly on compressed data, 
they maintain higher robustness toward multiple processing operations and even outperform the SBIs method, whose overall score even decreases by 0.66\% instead. On the other hand, none of the three methods can properly classify video deepfakes processed by heavy compression, resolution reduction, or video noise.

In addition to benchmarking overall performance, the assessment framework also provides the means to analyze the behavior of a method under one specific realistic situation and help reveal the mechanism behind it. For instance, it is interesting to observe that, regardless of the training data, the SBIs method is more robust to geometric transformation than the other two and retains a good ability to accurately classify a vertically flipped video. It is because the SBIs method is based on local forgery traces instead of the global inconsistency on the face. 

While the generalization problem is well-explored by synthetic data-based methods, how to improve robustness toward processing operations and distortions which exist in the real world is still an open question. This paper provides a systematic benchmarking approach that helps reveal the drawbacks of general deepfake detectors.
For instance, although the SBIs method demonstrates a good generalization ability in cross-dataset experiments in their paper \cite{Shiohara_2022_CVPR}, our assessment framework shows that it is susceptible to some common perturbations in the real world, such as video compression, video noise, and low resolution. 



\begin{figure}[h]
	\centering
    \includegraphics[width=0.95\linewidth]{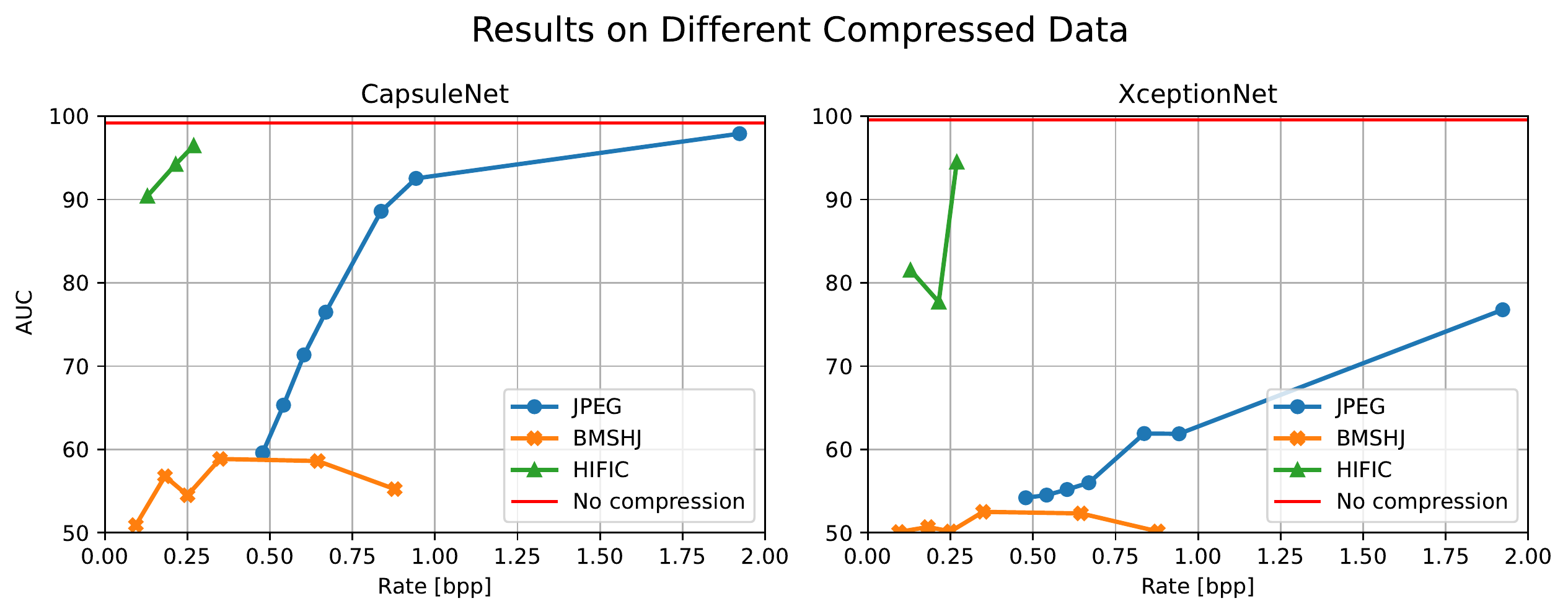}
	\caption{Detection performance on data compressed by conventional and AI-based coding algorithms.}
	\label{fig:codec}
\end{figure}


\subsection{Impact of Different Image Coding Algorithms}

The assessment framework additionally provides means to measure the impact of a specific type of processing operation on the performance of a deepfake detector. For instance, image compression operation is almost inevitable during the distribution of a fake image. Meanwhile, AI-based compression technologies have become increasingly popular and are often capable of obtaining relatively smaller bitstreams. However, it is unknown to which extent the learning-based compression algorithms will affect the deepfake detection methods compared to conventional JPEG compression.

In this subsection, a detailed comparison has been made between JPEG compression and two popular AI-based image compression methods, denoted by \textit{bmshj} \cite{balle2018variational} and \textit{hific} \cite{mentzer2020high} respectively. In detail, the Capsule-Forensics and XceptionNet methods are first trained on uncompressed data. Afterward, their performance on different compressed data is evaluated using the framework and is then reported in Figure \ref{fig:codec}.
As a result, the image compression operation generally brings more negative impact to XceptionNet than to the Capsule-Forensics method. The latter obtains relatively high AUC scores when the test data is compressed by JPEG with high compression factors.  
Although the \textit{bmshj}-based compression method is capable of achieving lower bitrates than JPEG, it brings significant negative impact to both detectors, whose predictions are close to random guess regardless of the select compression factor. 
On the contrary, both tested detectors are more robust to test data compressed using \textit{hific} codec than using JPEG operation or \textit{bmshj} codec, even with extremely low bitrates. The results reported in this section imply that \textit{hific} codec introduces fewer adversarial artifacts, which can interrupt the functionality of other AI-based detectors.

\section{Stochastic Degradation-based Augmentation}
To improve the ability of deepfake detection methods to handle realistic distortions and pre-processing operations, an effective data augmentation approach is proposed which leads to a robustness improvement. 
\subsection{Proposed Augmentation Method}

Standard data augmentation methods often introduce geometric and color space transformation to enrich training data and improve the model generalization ability. But according to our experiments, this type of augmentation technique is less effective for deepfake detection under realistic conditions. 

Motivated by a typical data acquisition and transmission pipeline in the real world, the stochastic degradation-based augmentation (SDAug) method is proposed.  
The main novelty of the proposed augmentation technique resides in the fact that it is driven by the typical operations that images and video are subject to in realistic conditions. 
Based on the observation of the data degradation process, a carefully designed augmentation chain is conceived, which allows the training data to better resemble real-world conditions and further boosts the performance of deepfake detection methods. 

Generally, the brightness and contrast of input image $x$ are first modified by image enhancement operator \textbf{\textit{enh}}. Afterward, the image is convoluted with an image blurring kernel \textbf{\textit{f}}, followed by additive Gaussian noise \textbf{\textit{n}}. In the end, \textbf{\textit{JPEG}} compression is applied to obtain the augmented training data $x_{\text{aug}}$. 
The augmentation chain is described by the following formula.
\begin{equation}
\centering
x_{\text{aug}}=\textbf{\textit{JPEG}}[(\textbf{\textit{enh}}(x)\circledast\textbf{\textit{f}})+\textbf{\textit{n}}]
\end{equation}

In addition, unlike the common data augmentation process, the SDAug method is implemented in a stochastic manner. The term `stochastic' can be interpreted in the following two aspects. Firstly, each aforementioned augmentation operation will occur with a certain probability in the augmentation chain. Secondly, each operation will use a random severity level for every frame. The realistic scenario is rather complex and does not necessarily consist of multiple types of distortions and processing operations. A random mixture of several distortions and severity levels can create more diversity in the augmented training data. Moreover, stochastic augmentation helps preserve more information from the original training data and therefore prevents accuracy loss on the high-quality data. 
In detail, the augmentation operations are explained in sequence as follows.

\textit{Enhancement}: The augmentation chain begins with an image enhancement operation. A probability of 50\% is adopted to apply either a brightness or a contrast operation on the training data which will be then non-linearly modified by a factor randomly selected from $[0.5, 1.5]$. 

\textit{Smoothing}: Image blurring operation is then applied with a selected probability of 50\%. Either Gaussian blur or Average blur filter is used with a kernel size varying in the range $[3, 15]$. 

\textit{Additive Gaussian Noise}: For each batch of training data, a probability of 30\% is adopted to add a Gaussian noise. The standard deviation of the Gaussian noise varies randomly in the interval $[0, 50]$.

\textit{JPEG Compression}: Finally, JPEG compression is applied with a selected probability of 70\%. The quality factor corresponding to the compression is randomly chosen in the range $[10, 95]$.

\begin{table}[t]
  \centering
  \caption{AUC (\%) scores of cores of the Capsule-Forensics, denoted as CapsuleNet, and XceptionNet methods tested on unaltered and distorted variants of FFpp. The suffix +DAug denotes that the model is trained with the proposed augmentation chain but without the stochastic manner. The suffix +SDAug denotes that the model is trained with the stochastic degradation-based augmentation technique. In this table, \textcolor[rgb]{ 1,  0,  0}{Red color} denotes the highest score and \textcolor[rgb]{ .267,  .447,  .769}{Blue color} denotes the second highest score.}
    \begin{adjustbox}{width=\textwidth}
     \begin{tabular}{clcccccccccccccc}
    \toprule
    \multirow{2}[4]{*}{Methods} & \multicolumn{1}{c}{\multirow{2}[4]{*}{TrainScheme}} & \multirow{2}[4]{*}{Unaltered} & \multicolumn{4}{c}{JPEG}      & \multicolumn{4}{c}{DL-Comp}   & \multicolumn{4}{c}{Gaussian Noise}  & \multirow{2}[4]{*}{\shortstack{Po-Gau\\ Noise}} \\
\cmidrule{4-15}          &       &       & 95    & 60    & 30    & AVG   & High  & Med   & Low   & AVG   & 5     & 10    & 30    & AVG   &  \\
    \midrule
    \midrule
    \multirow{5}[2]{*}{CapsuleNet} & \multicolumn{1}{c}{FFpp-Raw} & \textcolor[rgb]{ 1,  0,  0}{99.20} & 97.91 & 76.48 & 59.60 & 78.00 & 55.24 & 54.50 & 50.92 & 53.55 & 61.80 & 59.73 & 51.26 & 57.60 & 55.63 \\
          & \multicolumn{1}{c}{FFpp-Full} & 94.52 & 94.95 & \textcolor[rgb]{ .267,  .447,  .769}{92.18} & 84.50 & 90.54 & 86.83 & 60.98 & 55.69 & 67.83 & 89.03 & 78.54 & 57.95 & 75.17 & 64.87 \\
          & \multicolumn{1}{c}{FFpp-Augmix} & \textcolor[rgb]{ .267,  .447,  .769}{98.66} & \textcolor[rgb]{ 1,  0,  0}{98.68} & 79.67 & 57.62 & 78.66 & 71.10 & 53.51 & 51.61 & 58.74 & 75.11 & 61.93 & 56.80 & 64.61 & 59.19 \\
          & \multicolumn{1}{c}{FFpp-DAug} & 93.06 & 92.90 & 91.24 & \textcolor[rgb]{ .267,  .447,  .769}{88.90} & \textcolor[rgb]{ .267,  .447,  .769}{91.01} & \textcolor[rgb]{ .267,  .447,  .769}{90.35} & \textcolor[rgb]{ 1,  0,  0}{81.96} & \textcolor[rgb]{ 1,  0,  0}{70.00} & \textcolor[rgb]{ 1,  0,  0}{80.77} & \textcolor[rgb]{ .267,  .447,  .769}{92.50} & \textcolor[rgb]{ .267,  .447,  .769}{88.78} & \textcolor[rgb]{ .267,  .447,  .769}{79.99} & \textcolor[rgb]{ .267,  .447,  .769}{87.09} & \textcolor[rgb]{ .267,  .447,  .769}{86.63} \\
          & \multicolumn{1}{c}{FFpp-SDAug} & 98.16 & \textcolor[rgb]{ .267,  .447,  .769}{97.97} & \textcolor[rgb]{ 1,  0,  0}{96.36} & \textcolor[rgb]{ 1,  0,  0}{94.08} & \textcolor[rgb]{ 1,  0,  0}{96.14} & \textcolor[rgb]{ 1,  0,  0}{93.81} & \textcolor[rgb]{ .267,  .447,  .769}{71.41} & \textcolor[rgb]{ .267,  .447,  .769}{59.74} & \textcolor[rgb]{ .267,  .447,  .769}{74.99} & \textcolor[rgb]{ 1,  0,  0}{97.05} & \textcolor[rgb]{ 1,  0,  0}{93.89} & \textcolor[rgb]{ 1,  0,  0}{83.51} & \textcolor[rgb]{ 1,  0,  0}{91.48} & \textcolor[rgb]{ 1,  0,  0}{87.06} \\
    \midrule
    \multirow{5}[1]{*}{XceptionNet} & \multicolumn{1}{c}{FFpp-Raw} & \textcolor[rgb]{ 1,  0,  0}{99.56} & 76.77 & 56.00 & 54.20 & 62.32 & 50.16 & 50.37 & 50.10 & 50.21 & 50.12 & 51.00 & 50.36 & 50.49 & 51.02 \\
          & \multicolumn{1}{c}{FFpp-Full} & \textcolor[rgb]{ .267,  .447,  .769}{99.02} & \textcolor[rgb]{ 1,  0,  0}{99.00} & \textcolor[rgb]{ .267,  .447,  .769}{94.78} & 87.86 & \textcolor[rgb]{ .267,  .447,  .769}{93.88} & \textcolor[rgb]{ .267,  .447,  .769}{94.36} & 54.88 & 55.78 & 68.34 & 59.00 & 55.09 & 54.08 & 56.06 & 51.43 \\
          & \multicolumn{1}{c}{FFpp-Augmix} & 99.15 & 87.12 & 63.38 & 59.58 & 70.03 & 77.86 & 62.07 & 55.76 & 65.23 & 77.37 & 68.45 & 56.99 & 67.60 & 62.00 \\
          & \multicolumn{1}{c}{FFpp-DAug} & 89.51 & 89.47 & 89.27 & \textcolor[rgb]{ .267,  .447,  .769}{89.00} & 89.25 & 89.49 & \textcolor[rgb]{ 1,  0,  0}{88.71} & \textcolor[rgb]{ 1,  0,  0}{86.16} & \textcolor[rgb]{ .267,  .447,  .769}{88.12} & \textcolor[rgb]{ .267,  .447,  .769}{89.43} & \textcolor[rgb]{ .267,  .447,  .769}{89.30} & \textcolor[rgb]{ .267,  .447,  .769}{88.22} & \textcolor[rgb]{ .267,  .447,  .769}{88.98} & \textcolor[rgb]{ .267,  .447,  .769}{88.97} \\
          & \multicolumn{1}{c}{FFpp-SDAug} & 98.44 & \textcolor[rgb]{ .267,  .447,  .769}{98.25} & \textcolor[rgb]{ 1,  0,  0}{97.36} & \textcolor[rgb]{ 1,  0,  0}{96.12} & \textcolor[rgb]{ 1,  0,  0}{97.24} & \textcolor[rgb]{ 1,  0,  0}{98.03} & \textcolor[rgb]{ .267,  .447,  .769}{87.76} & \textcolor[rgb]{ .267,  .447,  .769}{82.74} & \textcolor[rgb]{ 1,  0,  0}{89.51} & \textcolor[rgb]{ 1,  0,  0}{97.37} & \textcolor[rgb]{ 1,  0,  0}{95.88} & \textcolor[rgb]{ 1,  0,  0}{91.71} & \textcolor[rgb]{ 1,  0,  0}{94.99} & \textcolor[rgb]{ 1,  0,  0}{94.57} \\
    \midrule
    \multirow{2}[3]{*}{Methods} & \multicolumn{1}{c}{\multirow{2}[3]{*}{TrainSet}} & \multicolumn{4}{c}{Gaussian Blur} & \multicolumn{5}{c}{Gamma Correction}  & \multicolumn{4}{c}{Resize}    & \multirow{2}[3]{*}{\shortstack{Overall\\Average}} \\
\cmidrule{3-15}          &       & 3     & 7     & 11    & AVG   & 0.1   & 0.75  & 1.3   & 2.5   & AVG   & x4    & x8    & x16   & AVG   &  \\
    \midrule
    \midrule
    \multirow{5}[2]{*}{CapsuleNet} & \multicolumn{1}{c}{FFpp-Raw} & 67.19 & 58.22 & 52.26 & 59.22 & 50.50 & \textcolor[rgb]{ 1,  0,  0}{98.86} & \textcolor[rgb]{ 1,  0,  0}{99.17} & \textcolor[rgb]{ .267,  .447,  .769}{96.12} & 86.16 & 67.48 & 52.18 & 53.10 & 57.59 & 65.35 \\
          & \multicolumn{1}{c}{FFpp-Full} & 85.72 & 58.83 & 56.05 & 66.87 & 56.02 & 93.86 & 93.87 & 85.44 & 82.30 & 84.65 & 65.34 & 52.02 & 67.34 & 75.01 \\
          & \multicolumn{1}{c}{FFpp-Augmix} & 90.86 & 54.08 & 50.67 & 65.20 & \textcolor[rgb]{ 1,  0,  0}{76.17} & \textcolor[rgb]{ .267,  .447,  .769}{98.57} & \textcolor[rgb]{ .267,  .447,  .769}{98.42} & 94.53 & \textcolor[rgb]{ 1,  0,  0}{91.92} & \textcolor[rgb]{ .267,  .447,  .769}{89.62} & 61.39 & 50.58 & 67.20 & 71.06 \\
          & \multicolumn{1}{c}{FFpp-DAug} & \textcolor[rgb]{ .267,  .447,  .769}{91.79} & \textcolor[rgb]{ .267,  .447,  .769}{86.00} & \textcolor[rgb]{ .267,  .447,  .769}{79.95} & \textcolor[rgb]{ .267,  .447,  .769}{85.91} & \textcolor[rgb]{ .267,  .447,  .769}{67.39} & 92.40 & 93.13 & 91.83 & 86.19 & 88.42 & \textcolor[rgb]{ .267,  .447,  .769}{77.06} & \textcolor[rgb]{ .267,  .447,  .769}{55.22} & \textcolor[rgb]{ .267,  .447,  .769}{73.57} & \textcolor[rgb]{ .267,  .447,  .769}{84.09} \\
          & FFpp-SDAug & \textcolor[rgb]{ 1,  0,  0}{96.86} & \textcolor[rgb]{ 1,  0,  0}{90.32} & \textcolor[rgb]{ 1,  0,  0}{80.31} & \textcolor[rgb]{ 1,  0,  0}{89.16} & 60.17 & 97.68 & 98.18 & \textcolor[rgb]{ 1,  0,  0}{96.91} & \textcolor[rgb]{ .267,  .447,  .769}{88.24} & \textcolor[rgb]{ 1,  0,  0}{93.54} & \textcolor[rgb]{ 1,  0,  0}{79.22} & \textcolor[rgb]{ 1,  0,  0}{58.05} & \textcolor[rgb]{ 1,  0,  0}{76.94} & \textcolor[rgb]{ 1,  0,  0}{86.16} \\
    \midrule
    \multirow{5}[2]{*}{XceptionNet} & \multicolumn{1}{c}{FFpp-Raw} & 68.76 & 55.61 & 50.70 & 58.36 & 54.66 & 98.66 & \textcolor[rgb]{ 1,  0,  0}{99.57} & 70.45 & 80.84 & 68.60 & 55.80 & 50.45 & 58.28 & 60.08 \\
          & \multicolumn{1}{c}{FFpp-Full} & \textcolor[rgb]{ .267,  .447,  .769}{96.36} & 70.51 & 54.50 & 73.79 & 51.38 & \textcolor[rgb]{ .267,  .447,  .769}{98.91} & \textcolor[rgb]{ .267,  .447,  .769}{98.84} & \textcolor[rgb]{ .267,  .447,  .769}{88.91} & 84.51 & \textcolor[rgb]{ .267,  .447,  .769}{93.47} & 75.30 & 60.55 & 76.44 & 75.50 \\
          & \multicolumn{1}{c}{FFpp-Augmix} & 90.45 & 62.58 & 53.00 & 68.68 & \textcolor[rgb]{ 1,  0,  0}{93.45} & \textcolor[rgb]{ 1,  0,  0}{99.33} & 98.32 & 85.87 & \textcolor[rgb]{ 1,  0,  0}{94.24} & 64.64 & 54.57 & 50.00 & 56.40 & 70.36 \\
          & \multicolumn{1}{c}{FFpp-DAug} & 89.22 & \textcolor[rgb]{ .267,  .447,  .769}{87.62} & \textcolor[rgb]{ .267,  .447,  .769}{85.28} & \textcolor[rgb]{ .267,  .447,  .769}{87.37} & 69.08 & 89.42 & 89.35 & 87.74 & 83.90 & 88.31 & \textcolor[rgb]{ .267,  .447,  .769}{81.30} & \textcolor[rgb]{ .267,  .447,  .769}{63.89} & \textcolor[rgb]{ .267,  .447,  .769}{77.83} & \textcolor[rgb]{ .267,  .447,  .769}{85.91} \\
          & FFpp-SDAug & \textcolor[rgb]{ 1,  0,  0}{98.31} & \textcolor[rgb]{ 1,  0,  0}{97.35} & \textcolor[rgb]{ 1,  0,  0}{94.51} & \textcolor[rgb]{ 1,  0,  0}{96.72} & \textcolor[rgb]{ .267,  .447,  .769}{80.48} & 98.25 & 98.44 & \textcolor[rgb]{ 1,  0,  0}{97.75} & \textcolor[rgb]{ .267,  .447,  .769}{93.73} & \textcolor[rgb]{ 1,  0,  0}{97.30} & \textcolor[rgb]{ 1,  0,  0}{86.26} & \textcolor[rgb]{ 1,  0,  0}{67.14} & \textcolor[rgb]{ 1,  0,  0}{83.57} & \textcolor[rgb]{ 1,  0,  0}{92.63} \\
    \bottomrule
    \end{tabular}%
   \end{adjustbox}
  \label{tab:SDAug-image}%
\end{table}%

\begin{table}[h]
  \centering
  \caption{AUC (\%) scores of three selected deepfake detection methods trained with the SDAug augmentation method on the distorted variants of the FFpp test set.}
  \begin{adjustbox}{width=\textwidth}
    \begin{tabular}{ccccccccccccccc}
    \toprule
    \multirow{2}[4]{*}{Methods} & \multirow{2}[4]{*}{TrainSet} & \multicolumn{2}{c}{Compression} & \multicolumn{2}{c}{Brightness} & \multirow{2}[4]{*}{Grayscale} & \multirow{2}[4]{*}{Contrast} & \multicolumn{2}{c}{Flipping} & \multicolumn{2}{c}{Resolution} & \multirow{2}[4]{*}{\shortstack{Gaussian \\ Noise}} & \multirow{2}[4]{*}{\shortstack{Vintage \\ Filter}} & \multirow{2}[4]{*}{Average} \\
\cmidrule{3-6}\cmidrule{9-12}          &       & C23 & C40 & Increase & Decrease &       &       & Horizontal & Vertical & x2    & x4    &       &       &  \\
    \midrule
    \midrule
    CapsuleNet & \multirow{6}[6]{*}{\shortstack{FFpp- \\ Raw}} & 77.97 & 54.14 & 73.31 & 70.62 & 68.38 & 69.31 & 73.13 & 63.20 & 65.43 & 56.99 & 54.14 & 72.94 & 66.63 \\
    +SDAug &       & 92.76 & 72.32 & 89.56 & 89.50 & 88.17 & 89.93 & 89.40 & 62.61 & 78.55 & 74.93 & 81.86 & 85.21 & 82.90 \\
\cmidrule{1-1}\cmidrule{3-15}    XceptionNet &       & 69.49 & 55.70 & 65.92 & 66.40 & 65.51 & 65.32 & 65.26 & 57.36 & 57.23 & 55.90 & 50.50 & 66.90 & 61.79 \\
    +SDAug &       & 94.89 & 80.53 & 93.36 & 93.05 & 92.64 & 92.98 & 92.32 & 57.65 & 88.42 & 81.60 & 88.91 & 90.47 & 87.24 \\
\cmidrule{1-1}\cmidrule{3-15}    SBIs   &       & 90.43 & 76.27 & 86.38 & 86.47 & 86.27 & 85.94 & 85.98 & 79.28 & 76.35 & 63.62 & 71.52 & 86.54 & 81.25 \\
    +SDAug &       & 89.31 & 76.60 & 85.55 & 86.24 & 85.41 & 84.76 & 85.48 & 78.67 & 77.06 & 64.31 & 77.13 & 86.11 & 81.39 \\
    \bottomrule
    \end{tabular}%
  \end{adjustbox}
  \label{tab:SDAug-video}%
\end{table}%



\begin{figure}[t]
	\centering
    \includegraphics[width=0.9\linewidth]{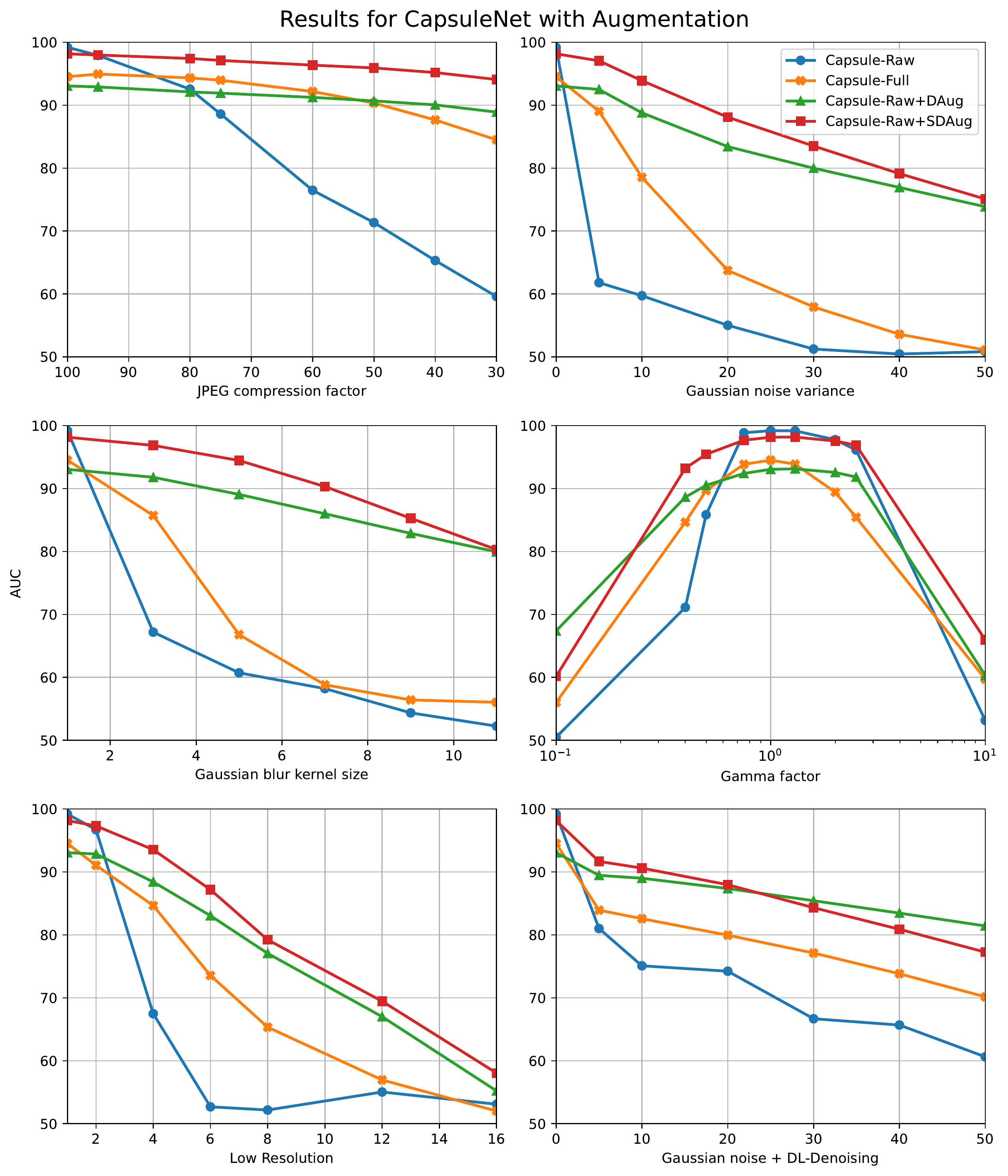}
	\caption{Performance comparison between models trained on FFpp-Raw only and trained with the proposed augmentation method.}
	\label{fig:SDAug-image}
\end{figure}


\begin{figure}[h]
	\centering
    \includegraphics[width=0.9\linewidth]{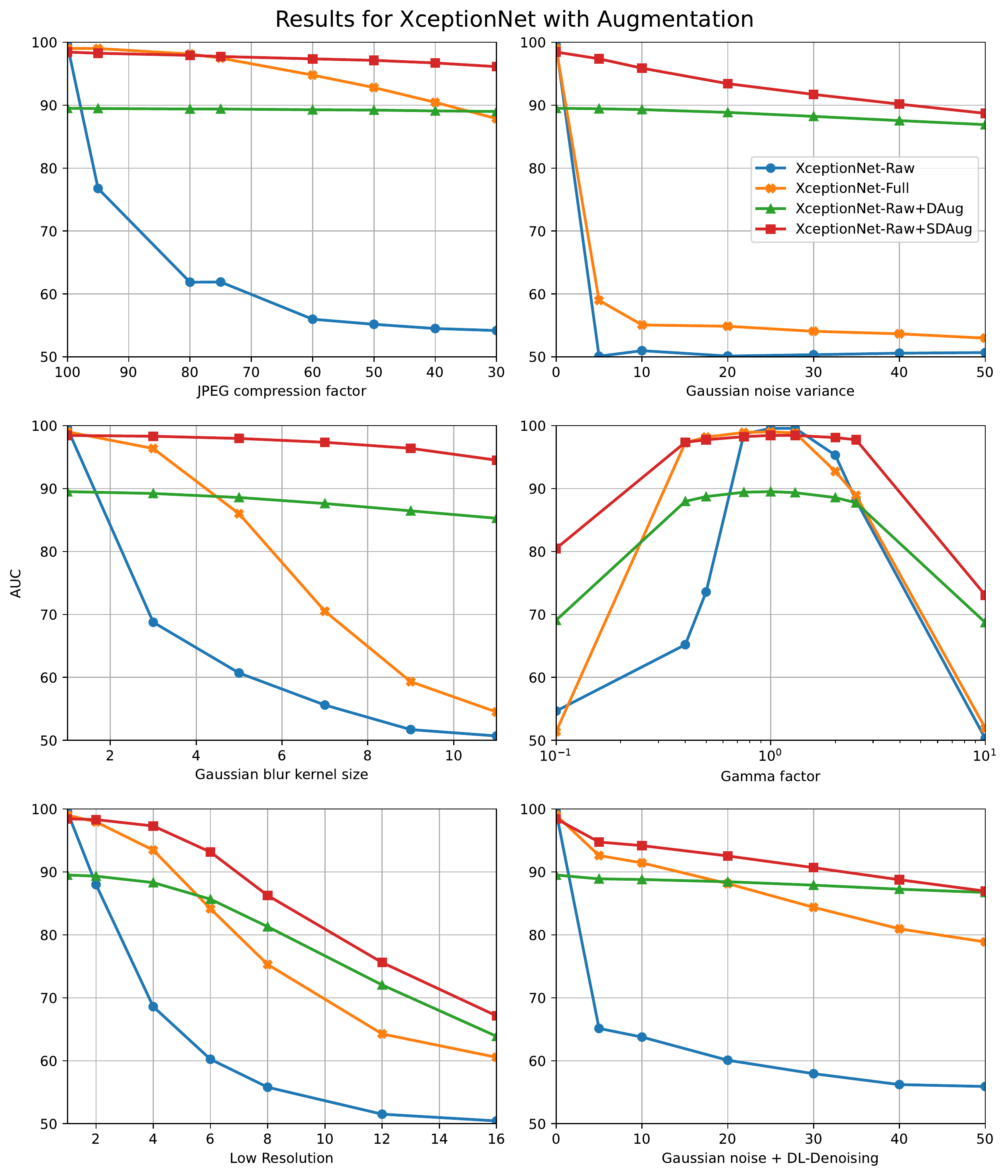}
	\caption{Performance comparison between models trained on FFpp-Raw only and trained with the proposed augmentation method.}
	\label{fig:SDAug-image2}
\end{figure}


\subsection{Experimental Results with Augmentation}

Table \ref{tab:SDAug-image} shows the evaluation results of the Capsule-Forensics and XceptionNet methods trained on the unaltered FFpp dataset together with the proposed augmentation strategy. The information regarding the models trained with the proposed stochastic degradation augmentation methods is denoted as \textbf{+SDAug}.

In comparison, it is evident that training with the stochastic degradation-based augmentation technique on the same dataset remarkably improves the performance on nearly all kinds of processed data even with intense severity. For example, previous experiments show that the detectors are more vulnerable to synthetic noises and blurry effects. The sub-figures in Figure \ref{fig:SDAug-image} and Figure \ref{fig:SDAug-image2} further illustrate the impact of increasing the severity of these distortions on the two detection methods. The data augmentation scheme significantly improves the robustness and meanwhile still maintains high performance on original unaltered data. 

It is worth noting that the performance improves not only on the four types of processing operations that appear during data augmentation but also on other different kinds of distortions. As shown in Table \ref{tab:SDAug-image} and the last two sub-figures in Figure \ref{fig:SDAug-image} and \ref{fig:SDAug-image2}, both detectors are much more robust towards learning-based compression, low-resolution effects, and other mixed distortions. A similar observation is obtained from the video deepfake assessment framework, see Table \ref{tab:SDAug-video}. Although these video processing operations are not present in the proposed augmentation chain, the SDAug technique brings performance improvement to the Capsule-Forensics and XceptionNet methods on nearly all kinds of processed video deepfakes.  

In order to compare with conventional augmentation methods based on geometric and color space transformation, the well-known Augmix \cite{hendrycks2019augmix} augmentation technique is evaluated under the same realistic assessment framework. This method generates multiple augmentation chains that work in parallel by randomly applying transformations to the training data. As a result, Augmix brings limited improvements to the robustness of the detector compared to SDAug, see Table \ref{tab:SDAug-image}. Its overall performance is even worse than simply training with low-quality data, which implies that the traditional data augmentation method is less practical when facing real-world distortions. 

Moreover, an ablation study is conducted in order to show the effectiveness of the stochastic mechanism. As shown in Table \ref{tab:SDAug-image}, although the degradation-based augmentation chain, denoted as DAug, is able to improve the performance on multiple processed data, the AUC score degrades heavily on the unaltered FFpp-Raw test set. The SDAug shows the most significant robustness improvement and meanwhile maintains high performance on original high-quality data.

\begin{table}[t]
  \centering
  \caption{Cross-dataset evaluation on Celeb-DFv2 (AUC(\%)) after training on FFpp dataset.}
    \begin{tabular}{ccccc}
    \toprule
    Deepfake Detector & Augmentation Method &       & FFpp  & Celeb-DFv2 \\
    \midrule
    \multirow{4}[2]{*}{Capsule} & No Aug &       & 99.20 & 54.39 \\
          & Augmix &      & 98.66      & 58.65 \\
          & DAug &        & 93.51      & 68.39 \\
          & SDAug &       & 97.82      & \textbf{71.86} \\
\cmidrule{1-2}\cmidrule{4-5}    \multirow{3}[2]{*}{XceptionNet} & No Aug &       & 99.56 & 50.00 \\
          & Augmix &      & 99.15      & 53.04 \\
          & DAug &        & 78.64      & 62.81 \\
          & SDAug &       & 98.44      & \textbf{73.88} \\
    \bottomrule
    \end{tabular}%
  \label{tab:cross}%
\end{table}%

Finally, a cross-dataset assessment has been conducted for the Capsule-Forensics and XceptionNet methods to evaluate the generalization ability of those models trained with the proposed augmentation technique. The results are shown in Table \ref{tab:cross}. The selected detectors are trained on the FFpp dataset but tested on the Celeb-DFv2 test set for frame-level AUC scores. The two methods obtain very low scores on the new dataset. On the contrary, the proposed augmentation scheme brings a noticeable performance improvement for both detectors on Celeb-DFv2, showing its capability to improve the generalization ability on unseen forensic face contents. 

\subsection{Limitations}
Although the proposed augmentation technique is in general very helpful in improving the robustness of deepfake detectors while facing various real-world image and video processing operations, some limitations have been observed from the previous results report, particularly in Table \ref{tab:SDAug-video}. First of all, the augmentation method provides limited help for the detection methods based on synthetic data, because it can possibly corrupt the manually designed forgery traces during training. Secondly, the proposed method shows less effect in the case of adverse geometric transformation, such as vertical flipping. 

\section{Conclusion}

Most of the current deepfake detection methods are designed to be as high performing as possible on specific benchmarks. But it has been shown that current assessment and ranking approaches employed in related benchmarks are less reliable and insightful. In this work, a more systematic performance assessment approach is proposed for deepfake detectors in realistic situations. To show the necessity and usage of the assessment framework, extensive experiments have been performed, where the robustness of three popular deepfake detectors is reported and analyzed. Furthermore, motivated by the assessment results, a new data augmentation chain based on a natural data degradation process has been conceived and shown to significantly improve the model's robustness against distortions from various image and video processing operations. The effectiveness and limitations of the proposed augmentation method have been also discussed in detail. 







\begin{backmatter}

\section*{Acknowledgements}
No additional acknowledgments.

\section*{Funding}
The authors acknowledge support from CHIST-ERA project XAIface (CHIST-ERA-19-XAI-011) with funding from the Swiss National Science Foundation (SNSF) under grant number 20CH21 195532.

\section*{Abbreviations}


\noindent
\begin{tabular}{l@{}cll}
AUC       &&Area Under Receiver Operating Characteristic Curve \\
DAug      &&Degradation-based Augmentation \\
DCNNs     &&Deep Convolutional Neural Networks \\
DFDC      &&Deepfake Detection Challenge \\
FFpp      &&FaceForensicsplusplus  \\
GANs      &&Generative Adversarial Networks  \\
SDAug     &&Stochastic Degradation-based Augmentation \\
TMC       &&Trusted Media Challenge  \\
\end{tabular}

\section*{Availability of data and materials}
The datasets used and/or analyzed during the current study are available from the corresponding author on reasonable request.

\section*{Ethics approval and consent to participate}
Not applicable

\section*{Competing interests}
The authors declare that they have no competing interests.

\section*{Consent for publication}
Not applicable

\section*{Authors' contributions}
All authors participated in the design of the analytics, performance measures, experiments, and writing of the manuscript. All authors read and approved the final manuscript.



\bibliographystyle{bmc-mathphys} 
\bibliography{bmc_article}      

\end{backmatter}
\end{document}